\documentclass[]{fairmeta}

\usepackage{subfigure}
\usepackage{graphicx}
\usepackage{amsthm}
\usepackage{amsmath}
\usepackage{bm}
\usepackage{bbding}

\usepackage{booktabs}

\usepackage{amsfonts}
\usepackage{float}
\usepackage{algorithm}
\usepackage{algorithmic}
\usepackage[normalem]{ulem}

\usepackage{colortbl}
\usepackage{xcolor}

\usepackage{multicol}
\usepackage{multirow}
\usepackage{enumerate}
\usepackage{enumitem}
\usepackage{listings}
\usepackage{tcolorbox}
\usepackage{wrapfig}
\usepackage{needspace}
\usepackage{booktabs}
\usepackage{nicefrac}
\usepackage{microtype}
\usepackage[table]{xcolor}
\usepackage{array}

\usepackage{wrapfig}
\usepackage{booktabs,threeparttable,siunitx,tabularx}
\title{\method: Orchestrating Cognitive Memory for Conversational Agents}

\author{Kai Zhang\textsuperscript{\dag}}
\author{Xinyuan Zhang\textsuperscript{\dag}}
\author{Hongda Jiang}
\author{Shiun-Zu Kuo}
\author{Hyokun Yun}
\author{Ejaz Ahmed}
\author{Shereen Oraby}
\author{Ziyun Li}
\author{Sanat Sharma}
\author{Ann Lee}
\author{Ahmed Aly}
\author{Anuj Kumar}
\author{Raffay Hamid}
\author{Xin Luna Dong\textsuperscript{\dag}}

\affiliation{Meta Reality Labs}

\abstract{Recall and personalization, enabled by memory, are pivotal to a lifelong conversational agent. However, simply expanding context windows with raw retrieval degrades reasoning quality, while training memory agents via standard reinforcement learning creates a severe credit assignment bottleneck in a multi-stage pipeline.  To solve this, we introduce \method, a framework that learns to manage and utilize a cognitively-structured memory—spanning user facts, preferences, and working memory. By introducing novel stage-wise process rewards and reward-decomposed contrastive refinement, \method\ provides isolated supervision for distinct memory operations (selective filtering, consolidation, and cue-driven recall) end-to-end. Across four datasets, SALIMORY records the highest Good Personalization rate at 39.8\% and the best QA accuracy at 72.9\%, while consistently producing the least vague and most factual memories, keeping vagueness under 20\% and factuality above 89\%.
}

\correspondence{\textsuperscript{\dag}\email{\{kkaizh,dylanz426,lunadong\}@meta.com}}

\metadata[Code]{\url{https://github.com/facebookresearch/SaliMory}}

\newcommand{\method}{\textsc{SaliMory}}

\begin{document}

\maketitle

\section{Introduction}
\label{sec:intro}
\begin{wrapfigure}{R}{0.65\linewidth}
    \centering
    \includegraphics[width=\linewidth]{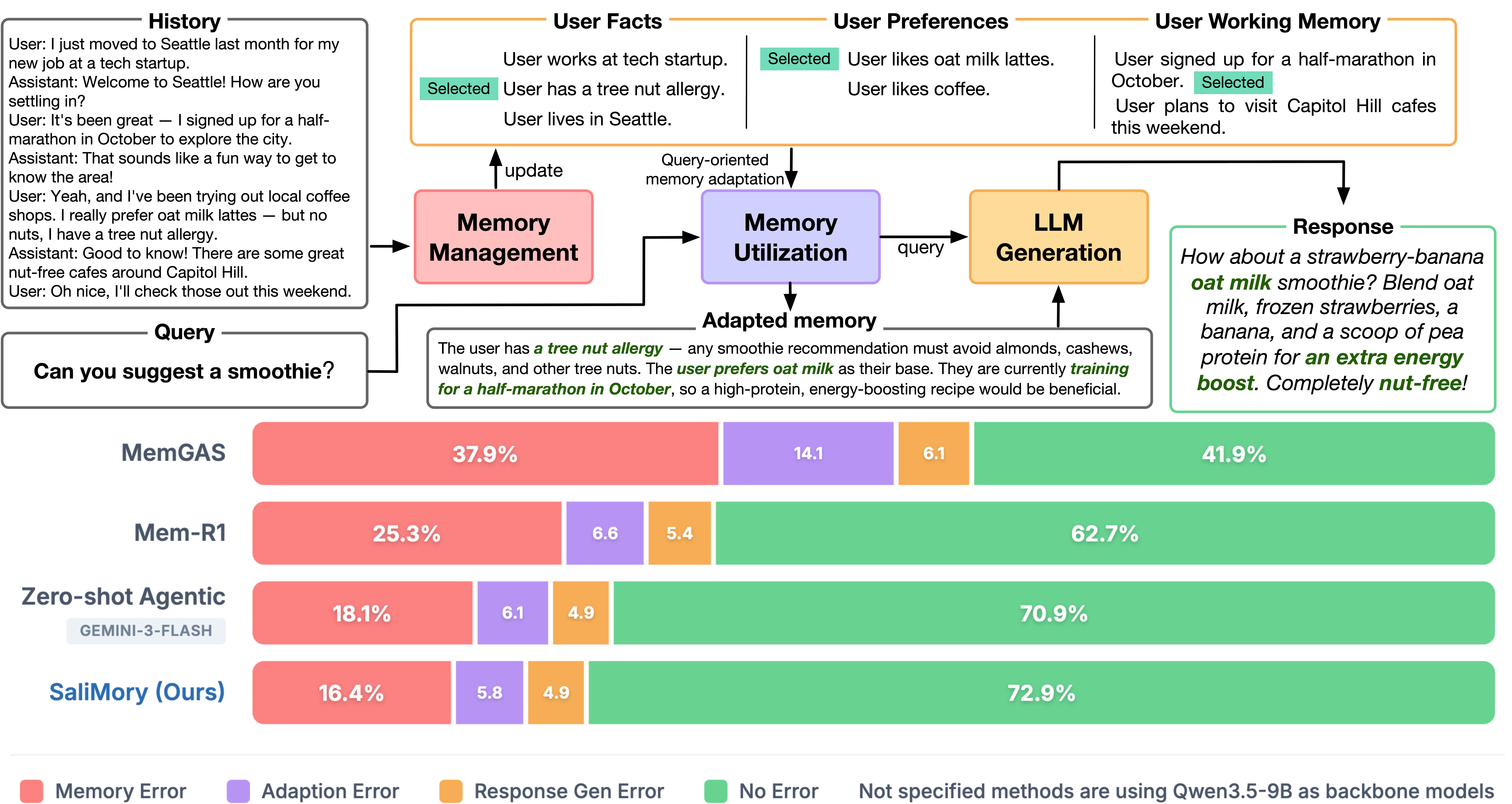}
    \caption{\textbf{Gated error attribution on LoCoMo.} We trace each incorrect response to its earliest point of failure in the pipeline (top). Memory errors---where salient information is lost or distorted during formation---dominate across baselines (e.g., 37.9\% for MemGAS), vastly exceeding profile generation and response generation errors combined.}
    \label{fig:error_analysis}
\end{wrapfigure}
The evolution of conversational AI is rapidly shifting from isolated, single-turn interactions toward lifelong digital companions. As AI agents become embedded in daily life, maintaining persistent context beyond the limits of standard LLM context windows has become a necessity~\citep{xu2022longtimenosee, bae2022keepmeupdated}. However, simply attaching an external memory storage is not enough: users' interaction histories are often vast, repetitive, and dominated by trivia. The real challenge is \emph{management}—deciding \textit{what to remember}, \textit{how to organize memories}, and \textit{what to recall for a given context}. Two main paradigms have emerged, but both fall short. One extreme is to store all past interactions and apply Retrieval-Augmented Generation (RAG) to surface relevant context when a new question arrives. As the memory accumulates, however, the retrieval results tend to become noisy and repetitive
~\citep{borgeaud2022, zhong2024memorybank, lee2024human}. Another extreme is to actively compress and curate long-term memories, but neither prompting-based~\citep{li2024ldagent, xu2025amem} nor reinforcement-learning–based~\citep{yan2025memoryr1, wang2025memalpha} variants have proved to maintain compressed memories to enable reliable capture of user preferences while preserving salient but nuanced details required for future conversations. Our error analysis on the LoCoMo benchmark~\citep{maharana2024locomo} reveals that in state-of-the-art solutions imprecise or incomplete memory can block well-personalized answers on up to 52\% of questions.

How do human beings cope with a lifetime of experiences without being overwhelmed? Cognitive science suggests we do not try to remember everything~\citep{atkinson1968, baddeley2000}: \emph{selective attention} filters what gets encoded, as recent traces remain active in \emph{working memory} to shape ongoing decisions~\citep{baddeleyhitch1974}.
What survives long-term is structurally organized: enduring \emph{facts} serve as hard constraints on what is true, while value-laden \emph{preferences} act as soft biases on what feels right~\citep{tulving1972}. Consolidation itself is shaped by downstream outcomes via dopaminergic reward~\citep{shohamy2010, lisman2011}.

Guided by these principles, we propose \textbf{\method}, an agentic memory-management framework built around three complementary stores: a \emph{factual snapshot} of verifiable facts about the user, a set of \emph{long-term preferences} capturing subjective tastes, and a \emph{short-term working memory} of recent details the user likely still has in mind. The factual snapshot supplies hard constraints; preferences supply soft criteria; working memory surfaces emerging interests worth revisiting.
A single memory-management module operates on these stores in three roles:
(i) \emph{selective attention}: judging which turns are salient enough to record; 
(ii) \emph{consolidation}: updating and forgetting memory; 
and (iii) \emph{cue-driven utilization}: retrieving and applying relevant memories at inference. 

Reinforcement learning (RL) for training a learnable memory has been extensively studied \citep{zhou2025mem1, zhang2026memrl, yan2025memoryr1}. However, existing reward designs are too sparse and distant to supervise intermediate memory decisions. We introduce two complementary mechanisms: The \textbf{stage-wise process reward} scores not only final-answer correctness and personalization, but also the quality of the resulting memory and the soundness of each intermediate decision (saliency, utilization). A \textbf{reward-decomposed contrastive refinement} step then amplifies the memory-management signal: within each Group Relative Policy Optimization (GRPO)~\citep{shao2024grpo} rollout batch, we construct role-specific preference pairs that upweight memory decisions in the overall objective. Together, \method\ enables a memory agent to memorize and utilize cognitive memory wisely in a human-like manner. In summary, this paper makes the following contributions.
\begin{itemize} 
    \item \textbf{Cognitively inspired memory architecture.} We separate factual snapshots, subjective preferences, and short-term working memory, and define three complementary memory-management roles---saliency filter, memory booster, and memory utilizer---that together ground long-horizon conversations. 
    \item \textbf{Memory-anchored RL.} We propose a process reward system that tracks memories in each stage and design reward-decomposed contrastive refinements over GRPO rollouts, assuring high-quality memory formation and end response generation.
    \item \textbf{New benchmark and evaluation protocol.} To fully evaluate memory's impact on conversational agents, we introduce a new \textbf{LoCoMo-P13n} benchmark with 2k personalizable queries. We also introduce a two-level evaluation protocol that jointly stresses end performance and intermediate memory quality. Empirically, a 9B \method\ achieves the highest Good Personalization rate and QA accuracy on four datasets, backed by the least vague and most factual memories throughout.
\end{itemize}
\section{Related Work}
\paragraph{Memory-Augmented LLM.}
Equipping LLMs with persistent context traditionally follows two paradigms.
Parametric Memory updates model weights via fine-tuning (e.g., FireAct~\citep{chen2023fireact}, AgentLumos~\citep{yin2024agent}) or soft parameters (SELF-PARAM~\citep{wang2024self}), but suffers from catastrophic forgetting and poor generalization to unseen queries. RAG-based Memory abstracts~\citep{borgeaud2022,lee2024human} experiences into external datastores, using semantic search to fetch relevant context.
However, it struggles to reliably distinguish critical objective facts and connections from transient remarks.
\vspace{-8pt}
\paragraph{Conversational Memory Structures.}
To overcome RAG's limitations, agentic systems organize context into distinct topologies. Linear memory treats history sequentially; MemGPT~\citep{packer2023memgpt} manages context using OS-like FIFO queues, while MemoryBank~\citep{zhong2024memorybank} decays older interactions.
Layered memory stratifies information to prioritize relevance, with MemoryOS~\citep{kang2025memory, li2025hello} segregating short, mid, and long-term stores based on access frequency and user's persona. Moreover, tree and graph-based memories map relational dependencies, with A-Mem~\citep{xu2025amem} constructing dynamic networks, AssoMem~\citep{zhang2025assomem} mimicking human associative memory for dense searches, and Mem0~\citep{chhikara2025mem0} utilizing entity-centric graphs. While these topologies improve organization, they rely on fragile heuristics and lack considerations on memory itself.
\vspace{-8pt}
\paragraph{Reinforcement Learning for Memory Management.}
Recent works increasingly formulate memory management as an RL problem.
MEM1~\citep{zhou2025mem1, zhang2025memgen} uses RL for learning to manage a memory, but its monolithic memory failed to take downstream task performance into consideration. Conversely, MemRL~\citep{zhang2026memrl} applies runtime RL to update retrieval Q-values but ignores memory creation quality. Memory-R1~\citep{yan2025memoryr1} and Mem-$\alpha$~\citep{wang2025memalpha} capture connections between the memory creation and downstream tasks, which train agents to execute explicit memory management and utilization. While pioneering, their monolithic rewards obscure where failures stem from in a multi-stage pipeline \citep{tan2026hindsight}.

Different from these previous efforts, \method\ connects downstream task performance and memory quality jointly and provides novel fine-grained credit assignments required to optimize a structured, multi-stage memory system end-to-end via RL algorithm.
\section{Methodology}
\label{sec:method}
\begin{figure}[t]
    \centering
    \includegraphics[width=\textwidth]{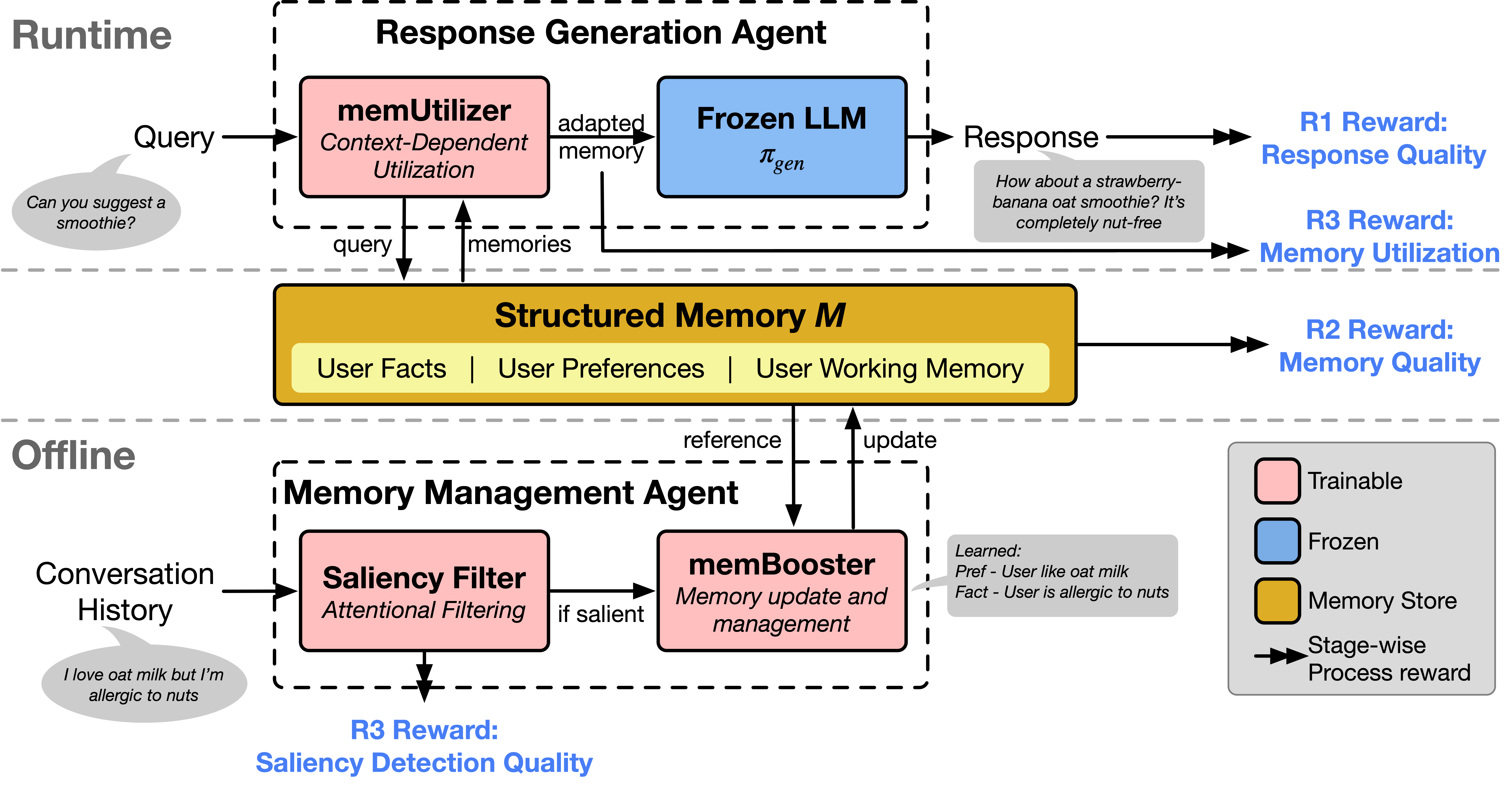}
    \caption{\method\ learns to manage and utilize cognitive memory for conversational agents.}
    \label{fig:architecture}
\end{figure}

We present \method, an end-to-end agentic memory framework designed to coordinate and optimize conversational memory for downstream response generation.

\subsection{Problem Definition and Memory Bank}
\label{sec:problem}
In \textit{long-horizon conversational} settings, an agent receives all conversation history $\mathcal{H}$ and a current user query $q$. The goal is to produce a {\em contextualized} and \textit{personalized} answer $\hat{a}$ that satisfies two criteria: accurately answering the query, and seamlessly grounding in the personal history.

Because $\mathcal{H}$ grows unboundedly and is dominated by trivia, we do not condition on it directly.
Instead, we maintain a dynamically evolving \textit{memory bank} $\mathcal{M}$, derived from $\mathcal{H}$, and reformulate the task as $\hat{a} = \pi_{\text{gen}}(q, \mathcal{M}).$ Inspired by human memory systems, $\mathcal{M}$ is not a flat buffer but three dedicated stores, $\mathcal{M} = (\bar F, \bar P, \bar W)$:
(i) \emph{factual bank $\bar F$}: objective facts that are verifiable (e.g., being lactose intolerant); 
(ii) \emph{preference bank $\bar P$}: subjective tastes and leanings that act as soft criteria for stylistic tailoring (e.g., preference for oat milk);
(iii) \emph{working memory $\bar W$}: recent salient turns the user likely still has in mind (e.g., asking about organic food this morning).
Separating long-term profile from short-term context, and objective facts from subjective preferences, allows the agent to differentiate, yielding sharper personalization than a flat context window allows.


Given this structured memory, the core challenge shifts from storage to \emph{dynamic management}. 
An effective memory manager must autonomously resolve three sequential decisions: 
\begin{itemize}
    \item[(i)] \textit{What enters memory?} Act as an attentional gate, deciding which turns carry salient information worth extracting and which to discard. 
    \item[(ii)] \emph{How is memory updated?} Route each extraction to the correct store ($\bar F$, $\bar P$, or $\bar W$), reconcile with existing entries, and retire stale items from working memory. 
    \item[(iii)] \emph{How is memory used?} At query time, synthesize relevant entries across the three stores into a query-adapted profile that conditions the generator.
\end{itemize}


\subsection{The \method\ Architecture and RL Framework}
\label{sec:e2eRL}

To separate the cognitive load of memory management from conversational reasoning for latency optimization,
the \method\ architecture (Figure \ref{fig:architecture}) 
contains two distinct computational stages: an \textit{offline stage} that builds and maintains the structured memory, and a \textit{runtime stage} that leverages it.

\textbf{The Offline Stage} processes each session of conversations asynchronously to produce an updated memory bank $\mathcal{M}$. First, the \textit{Saliency Filter} drops transient noise. Second, the \textit{memBooster} processes salient turns, incrementally updates the factual bank $\bar F$ and the preference bank $\bar P$, and maintains the working memory bank $\bar W$ within a sliding time window.   

\textbf{The Runtime Stage} executes when a new user query $q$ arrives.
The \textit{memUtilizer} finds relevant memories from the current memory bank, obtaining an adapted user profile $\mathcal{M}_q$. A separate frozen LLM ($\pi_{gen}$) then produces the final response $\hat{a} = \pi_{gen}(q, \mathcal{M}_q)$.

To autonomously execute the three memory decisions defined above, \method\ trains a single, unified policy model $\pi_\theta$ via RL rather than relying on separate specialized models.
By conditioning $\pi_\theta$ on distinct instruction prompts, this single model learns to act as the saliency filter, the memory booster, and the memory utilizer.

We formally define the \textit{state} at any given step as the combination of the conversational context and the memory bank $(q, \mathcal{H}, \mathcal{M})$. 
The \textit{action space} corresponds to the generative outputs of our policy model $\pi_\theta$ , including the saliency decision, the updated memory ${\cal M}$, and the query-dependent profile $\mathcal{M}_q$.
Our goal for the RL training is to  manipulate the state through these actions to maximize the quality of the final downstream response $\hat{a}$.

\subsection{Stage-Wise Process Reward System}
\label{sec:reward}

Recent advances in RL for reasoning models \citep{guo2025deepseekr1} rely heavily on Reinforcement Learning with Variable Reward (RLVR), where the model is optimized purely on a final, outcome-based reward.
However, applying pure RLVR to a complex, multi-stage memory pipeline creates a severe credit assignment bottleneck.
If the frozen generator produces an incorrect or poorly personalized final answer, an outcome-only reward cannot determine where the pipeline failed.

To solve this, \method\ decomposes the learning signals into a dense, hierarchical stage-wise process reward, providing targeted, orthogonal feedback to each specific role using LLM judges.

\textbf{Reward 1: Response Quality (Multiplicative Gating)}
The primary objective evaluates the final generated answer $\hat{a}$ against the ground truth $a$ for accuracy $\alpha \in \{0,1\}$ and personalization quality $p \in [0,1] \cup \{\texttt{n/a}\}$.
We formulate this reward using a strict multiplicative gate:
\begin{equation}
\label{eq:r1}
R_1 = \begin{cases}
\alpha \cdot (1 + \lambda_p \cdot p) & \text{if personalization is applicable} \\
\alpha & \text{otherwise}
\end{cases}
\end{equation}
This multiplicative form embeds a critical inductive bias: personalization is only valuable when the underlying answer is factually correct.
By multiplying the terms, an incorrect but highly personalized response receives $R_1 = 0$, preventing the policy from learning to "decorate" hallucinated answers.

\textbf{Reward 2: Memory Quality (Asymmetric Penalty)}
To directly supervise the memory booster's writing behavior independently of the final answer, a judge evaluates the finalized memory bank $\mathcal{M}_K$ for factuality $\phi_f$ (fraction of entries faithful to the source) and vagueness $\phi_v$ (fraction of entries losing actionable specificity).
Defining the composite memory quality as $\gamma = \phi_f \cdot (1 - \phi_v)$, we apply an asymmetric reward:
\begin{equation}
\label{eq:r2}
R_2 = -\lambda_{\text{pen}} \cdot (1 - \gamma) + \lambda_{\text{bon}} \cdot \gamma, \qquad \text{where } \lambda_{\text{pen}} > \lambda_{\text{bon}}
\end{equation}
The asymmetry $\lambda_{\text{pen}} = 2 \times \lambda_{\text{bon}}$ reflects the fact that low-quality memories actively sabotage downstream generation.
This asymmetric penalty exerts a conservative pressure on the booster: it learns that it is strictly better to omit a marginal memory than to encode a vague one.

\textbf{Reward 3: Utilization and Filter Quality}
The final level provides targeted, isolated feedback to the utilizer and the saliency filter.
It is defined as:
\begin{equation}
\label{eq:r3}
R_3 = \lambda_{\mu} \cdot \mu + \lambda_{\kappa} \cdot \kappa
\end{equation}
Here, $\mu \in [0,1]$ evaluates how effectively the adapted profile $\mathcal{M}_q$ captures question-relevant constraints. Meanwhile, $\kappa$ evaluates the saliency filter’s binary gating decisions against ground-truth labels using a recall-weighted blend: $\kappa = (1 - w_r) \cdot \text{Prec} + w_r \cdot \text{Rec}$ because missing a salient turn (false negative) is worse than flagging an extra non-salient one (false positive).

\textbf{Total Reward}
The total per-episode reward $R_{total} = R_1 + R_2 + R_3$.
The key property is \emph{role-specific credit assignment}: $R_1$ (range $[0, 1{+}\lambda_p]$) dominates as the primary objective, $R_2$ (range $[-\lambda_{\text{pen}}, \lambda_{\text{bon}}]$) directly shapes the booster's writing behavior independently of response quality, and $R_3$ (range $[0, \lambda_\mu{+}\lambda_\kappa]$) separately shapes the utilizer's synthesis and the filter's gating.

\subsection{Reward-Decomposed Contrastive Learning}
\label{sec:contrastive}
Standard GRPO assigns a single advantage scalar to all generative traces within a rollout---affecting the filter, booster, and utilizer equally even though we use role-specific process rewards.
Consequently, a rollout might achieve a high total reward simply due to a fortunate generation by the utilizer, which inadvertently sends the upstream booster an unearned positive gradient even if it writes vague or poor memories.
To fully resolve this intra-episode credit assignment bottleneck, \method\ exploits our reward decomposition to build {\em role-specific contrastive preference pairs}, attributing credit to each role from its own contribution.

\textbf{Booster Contrastive Pairs ($\mathcal{L}_{\text{cont}}^{\text{boost}}$)}
For the memory booster, we select winner ($w$) and loser ($l$) rollouts based strictly on the $R_2$ memory quality score, completely ignoring the total episode reward.
Let $\mathcal{P}_b$ be the set of paired booster traces: at each turn where the two rollouts produce different memory writes, the winner's output $y^w_j$ is paired with the loser's $y^l_j$ under the same booster prompt $p^b_j$.
We apply a DPO-style contrastive loss over these pairs, scaled by a gap weight $\Delta_{R_2} = |R_2^w - R_2^l|$:
\begin{equation}
\label{eq:boost_cont}
\mathcal{L}_{\text{cont}}^{\text{boost}} = \frac{1}{|\mathcal{P}_b|} \sum_{j \in \mathcal{P}_b} \Delta_{R_2} \cdot \Big[-\log \sigma\Big(\beta \big(\log \pi_\theta(y^w_j \mid p^b_j) - \log \pi_\theta(y^l_j \mid p^b_j)\big)\Big)\Big]
\end{equation}
The gap weight modulates the gradient magnitude by preference confidence: when the quality gap between two memory writes is large, the contrastive signal strongly pushes the booster; when marginal, the loss contributes minimally, avoiding noisy updates from ambiguous comparisons.

\textbf{Utilizer Contrastive Pairs \& Similarity Gating ($\mathcal{L}_{\text{cont}}^{\text{util}}$)}
For the utilizer, winner and loser rollouts are selected based on the utilization score $\mu$ (from $R_3$).
However, a naive comparison here is causally confounded: a superior adapted profile might simply be the result of the booster providing better upstream memories to work with.
To isolate the utilizer's true synthesis ability from upstream memory noise, we introduce a gating mechanism based on the Jaccard similarity between the memory banks of the two rollouts:
\begin{equation}
\label{eq:mem_sim}
\text{sim}(\mathcal{M}^w, \mathcal{M}^l) = \frac{|\text{entries}(\mathcal{M}^w) \cap \text{entries}(\mathcal{M}^l)|}{|\text{entries}(\mathcal{M}^w) \cup \text{entries}(\mathcal{M}^l)|} > \tau_{\text{sim}}
\end{equation}
Contrastive pairs are only formed if this similarity exceeds the threshold $\tau_{\text{sim}}$.
This strict similarity gate ensures that any difference in profile quality reflects the utilizer's own synthesis ability---how it selected, prioritized, and phrased the context---rather than differences in the underlying available information.
Below-threshold pairs are discarded, so the utilizer only receives gradients it actually earned.
With $\mathcal{P}_u$ denoting the admitted profile pairs and $\Delta_{\mu} = |\mu^w - \mu^l|$, the utilizer contrastive loss is:
\begin{equation}
\label{eq:util_cont}
\mathcal{L}_{\text{cont}}^{\text{util}} = \frac{1}{|\mathcal{P}_u|} \sum_{(w,l) \in \mathcal{P}_u} \Delta_{\mu} \cdot \Big[-\log \sigma\Big(\beta \big(\log \pi_\theta(\rho^w \mid p^u) - \log \pi_\theta(\rho^l \mid p^u)\big)\Big)\Big]
\end{equation}
where $\rho$ is the adapted profile and $p^u$ the utilizer prompt.

These two role-specific contrastive refinements combine to form our additional learning boost:
\begin{equation}
\label{eq:contrast}
\mathcal{L}_{\text{Cont}} = \lambda_b \cdot \mathcal{L}_{\text{cont}}^{\text{boost}} \;+\; \lambda_u \cdot \mathcal{L}_{\text{cont}}^{\text{util}}
\end{equation}
Ultimately, this boost does not replace the main RL objective, but acts alongside it to provide the precise credit assignment necessary to train a highly structured cognitive pipeline.

\subsection{Full Objective and Policy Optimization}
\label{sec:optimization}
The complete per-step loss seamlessly integrates our stage-wise process rewards into the standard GRPO~\citep{shao2024grpo} objective with our novel contrastive boost:
\begin{equation}
\label{eq:full_loss}
\mathcal{L}_{\text{total}} = \frac{1}{N}\sum_{\xi} \mathcal{L}_{\text{GRPO}}(\xi) \;+\; \mathcal{L}_{\text{Cont}}
\end{equation}
where $N$ is the total number of generative traces and $\mathcal{L}_{\text{GRPO}}$ is the standard clipped surrogate objective with KL regularization optimized using our hierarchical rewards $R_{total}$.

Crucially, this combined objective finalizes our solution to the credit assignment problem through selective gradient routing.
The main GRPO loss $\mathcal{L}_{\text{GRPO}}$ provides the baseline update and trains the saliency filter, whereas $\mathcal{L}_{\text{Cont}}$ backpropagates only through the booster- and utilizer-conditioned traces of the shared policy, based on their isolated preference pairs.

\section{Experimental Setup}
\label{sec:experiments_setup}
\textbf{Dataset.}
To evaluate \method, the dataset must satisfy Chat AI scenarios where user conversation history, current query and gold response are provided. Thus, we select  LoCoMo~\citep{maharana2024locomo}, a long-term conversational memory benchmark with 10 users and ${\sim}$58 multi-session dialogues per user. The original benchmark provides query types across five categories (multi-hop, temporal, open-domain, single-hop, adversarial) which are all memory recall queries. We argue that personalized engagements are crucial for conversational AI, yet they are absent from existing memory datasets such as LoCoMo and LongMemEval \citep{wu2024longmemeval}. To address this, we extend LoCoMo and introduce a new dataset named {\bf LoCoMo-P13n} where two personalizable query categories: \emph{recommendation} and \emph{implicit personalization} are curated to support such research (more details can be seen in Appendix \ref{appendix:dataset_details}). Furthermore, to validate the generalizability of \method, we also evaluate on an internal dataset where real-world Chat AI traffic is collected as in Section \ref{sec:efficiency_generalizability_analysis}. 

\textbf{Models and training.}
The trainable policy $\pi_\theta$ uses Qwen3.5-9B-Instruct\citep{team2026qwen3}; the frozen generator $\pi_{\text{gen}}$ uses Qwen3-235B-A22B-Instruct\citep{yang2025qwen3} and the judge uses GPT-4o\citep{hurst2024gpt}. For fair comparisons, we reproduce all baselines with the same settings; for zero-shot agentic method, we use Gemini-3-flash~\citep{pichai2025new} to play as the agent to do memory management and utilization. Full training details and hyperparameters setup are in Section \ref{appendix:hyperparams} \footnote{We also provide a scale of backbone models and training strategy ablation analysis in Appendix \ref{appendix:supplementary_analysis}.}.

\textbf{Baselines and ablations.}
To validate the superiority of \method\ compared to SOTA memory studies, we compare against four existing distinct popular memory paradigms: \textbf{Infinite Context}, which treats the entire history as memory; \textbf{RAG} -- A-Mem\citep{xu2025mem}, MemoryGAS\citep{xu2025towards}, which process history into a memory base for future use; \textbf{Zero-shot Agentic}, which uses powerful LLM as agent for memory operation and \textbf{Learnable Memory} -- Mem-R1\citep{yan2025memoryr1}, which learns to memorize and utilize user info in an RL framework. 

\textbf{Evaluation Protocol.}
To comprehensively assess memory-driven conversational agents, we introduce a novel LLM-as-a-Judge protocol that thoroughly evaluates fine-grained personalization and memory bottlenecks, 
with full implementation details provided in Appendices \ref{appendix:p13n_protocol} and \ref{appendix:metric_details}.
\begin{itemize}
    \item End-to-End Personalization Quality: Standard memory QA accuracy~\citep{jiang2025memory} fails to capture whether an agent effectively uses memory.
Thus, we introduce a \textit{Three-Step Personalization Judge}.
Step 1 determines if the response incorporates user memories.
Step 2 classifies personalized responses as \textit{Good} (rich and value-adding), \textit{Basic} (superficial), or \textit{Bad} (intrusive, irrelevant, or hallucinated).
Step 3 evaluates unpersonalized responses to determine if relevant memories existed but were ignored (\textit{Underuse}), or there is genuinely no personalization potential (\textit{NA}).
    \item Intermediate Memory Eval: We evaluate the finalized memory bank for \textit{Vagueness} (entries lacking actionable specificity) and \textit{Factuality} (faithfulness to the source).
We also measure \textit{Utilization}, scoring how well the synthesized profile captures question-relevant constraints.
    \item Gated Error-Bucket: For every incorrect response, we trace the failure to its origin—Memory Error, Profile Error, or Generation Error—to pinpoint specific reasoning bottlenecks.
\end{itemize}
\section{Experimental Results and Analysis}
\label{sec:experimental_results}
In this section, we will answer \textbf{RQ 1.} in Section \ref{sec:comparative}, \textbf{RQ 2.} in Section \ref{sec:ablation}, \ref{sec:training_dynamics} and \textbf{RQ 3.} in Section \ref{sec:efficiency_generalizability_analysis}, \ref{sec:error_analysis}, supplementary results can be seen in Appendix \ref{appendix:supplementary_analysis}:
\begin{itemize}
    \item \textbf{RQ 1.} Does \method\ exhibit better performances over existing memory approaches?
    \item \textbf{RQ 2.} What contributes to the improvements made by \method?
    \item \textbf{RQ 3.} How does \method\ generalize to different query types in Chat AI scenarios?
\end{itemize}

\subsection{Comparative Study}
\label{sec:comparative}
\begin{table}[t]
\centering
\small
\caption{\textbf{Main results on LoCoMo and LongMemEval.} Best in \textbf{bold}. $\uparrow$/$\downarrow$: higher/lower is better.}
\label{tab:main_results}
\resizebox{\textwidth}{!}{%
\begin{tabular}{@{}l >{\columncolor{orange!15}}c >{\columncolor{orange!15}}c cccc cc@{}}
\toprule
& \multicolumn{6}{c}{\textbf{Response Quality}} & \multicolumn{2}{c}{\textbf{Memory Quality}} \\
\cmidrule(lr){2-7}
& QA & \multicolumn{5}{c}{Personalization Categories (\%)} & \multicolumn{2}{c}{} \\
\cmidrule(lr){2-2} \cmidrule(lr){3-7} \cmidrule(lr){8-9}
\textbf{Method} & Accuracy\ $\uparrow$ & Good $\uparrow$ & Basic & Bad $\downarrow$ & Underuse\ $\downarrow$ & NA\ $\downarrow$ & Vague $\downarrow$ & Factuality\ $\uparrow$ \\
\midrule
\multicolumn{9}{l}{(a) Comparative Results on \textbf{LoCoMo and LoCoMo-P13n}} \\
Infinite Context & 33.4 & 18.2 & 13.6 & \textbf{0.8} & 39.3 & 27.3 & - & - \\
Zero-shot Agentic & 70.9 & 6.6 & 20.9 & 6.1 & 42.8 & 23.6 & 29.7 & 86.1 \\
A-Mem & 61.6 & 21.9 & \textbf{49.1} & 6.4 & 8.5 & 13.1 & 44.2 & 44.3 \\
MemoryGAS & 49.1 & 5.1 & 8.2 & 11.6 & 8.9 & 66.2 & 46.7 & 59.3 \\
Mem1 & 15.2 & 1.6 & 7.9 & 19.6 & 17.1 & 53.8 & 59.3 & 31.6 \\
Mem-$\alpha$ & 46.1 & 16.0 & 31.4 & 6.4 & 9.5 & 36.7 & 36.2 & 72.6 \\
Mem-R1 & 62.7 & 16.3 & 21.8 & 7.4 & 19.9 & 34.6 & 35.7 & 76.8 \\
\midrule
\textbf{\method} & \textbf{72.9} & \textbf{39.8} & 44.1 & 1.4 & \textbf{5.3} & \textbf{9.4} & \textbf{19.6} & \textbf{93.9} \\
\midrule
\midrule
\multicolumn{9}{l}{(b) Comparative Results on \textbf{LongMemEval-m}, results on \textbf{LongMemEval-s} are in Appendix \ref{appendix:supplementary_analysis}} \\
Infinite Context & 17.3 & 4.6 & 19.1 & 1.3 & 42.6 & 32.4 & - & - \\
Zero-shot Agentic & 62.2 & 13.6 & 22.2 & 2.8 & 11.2 & 50.2 & 9.6 & 79.6 \\
A-Mem & 49.4 & 12.2 & 21.2 & 2.0 & 11.4 & 53.2 & 27.7 & 45.6 \\
MemoryGAS & 51.2 & 7.3 & 19.4 & 2.1 & 15.2 & 56.0 & 21.9 & 57.3 \\
Mem1 & 50.4 & 8.0 & 38.8 & 1.8 & 13.8 & 37.6 & 11.6 & 56.0  \\
Mem-$\alpha$ & 57.6 & 14.6 & 34.4 & 2.4 & 15.8 & 32.8 & 9.5 & 80.2  \\
Mem-R1 & 61.5 & 14.1 & 37.6 & 3.9 & 27.1 & 17.3 & 9.9 & 81.9 \\
\midrule
\textbf{\method} & \textbf{66.4} & \textbf{22.6} & \textbf{39.3} & \textbf{1.3} & \textbf{6.2} & \textbf{30.1} & \textbf{9.3} & \textbf{89.7} \\
\bottomrule
\end{tabular}
}
\end{table}

\textbf{End-to-End Quality.}
As presented in Table \ref{tab:main_results}, \method\ achieves the highest E2E Accuracy (72.9\%) and reaches 39.8\% Good Personalization rate on recall and personalizable queries, respectively.
Crucially, despite that \method\ being trained on a compact backbone model (Qwen3.5-9B), it outperforms the Zero-shot baseline powered by a much larger model Gemini-3-Flash—delivering a 33.2-point absolute gain in Good Personalization.
Notably, A-Mem achieves a higher Basic personalization rate than \method, but its Good personalization rate (21.9\%) is much lower due to low memory quality.
Furthermore, SALIMORY nearly eradicates the Bad Personalization rate (1.4\% vs. Mem-R1's 7.4\%), proving our R1 reward stops the model from `decorating' hallucinated responses.

\textit{Why do prior memory methods fail at scale?} Different paradigms prioritize the wrong stages. Infinite Context provides all information but fails at utilization (39.3\% underuse) because raw logs overwhelm the reasoning capacity. Conversely, methods like MemGAS and Mem-R1 focus heavily on memory formation but struggle to effectively retrieve and apply it, resulting in high NA rates (66.2\% and 34.6\%). \textit{Why does \method\ help?} By orchestrating memory into cognitively-structured stores and applying a role-decomposed RL framework, \method\ explicitly co-optimizes both memory formation and cue-driven utilization, drastically reducing NA (9.4\%) and underused (5.3\%) responses.

\textbf{Memory Quality.} \method\ reduces the Vague memory rate to 19.6\% (vs. 35.7\% for Mem-R1) and increases the Factual rate to 93.9\%. This validates that \method\ successfully filters noise during memory formation. While the Gemini-3-Flash agentic baseline achieves decent memory factuality (86.1\%) and high QA Accuracy (70.9\%), its dismal personalization score proves that without dedicated training for memory utilization, even massive frontier models completely waste memory context. Following our stage-wise process reward and contrastive refinement, \method\ bridges this gap, enabling a significantly smaller open-source model to effectively couple memory formation with downstream utilization.

\subsection{Ablation Study}
\label{sec:ablation}

\begin{table*}[t]
\centering
\setlength{\tabcolsep}{5pt}
\renewcommand{\arraystretch}{0.95}
\caption{\textbf{RL training ablation studies.} Best in each section in \textbf{bold}. $\uparrow$/$\downarrow$: higher/lower is better.}
\label{tab:ablations}
\resizebox{\textwidth}{!}{%
\begin{tabular}{@{}l>{\columncolor{orange!15}}c >{\columncolor{orange!15}}c cc ccc@{}}
\toprule
& \textbf{QA} & \multicolumn{3}{c}{\textbf{Personalization Quality}} & \multicolumn{2}{c}{\textbf{Memory Quality}} & \textbf{Memory} \\
\cmidrule(lr){3-5} \cmidrule(lr){6-7}
\textbf{Config} & Accuracy $\uparrow$ & Good$\uparrow$ & Basic & Bad$\downarrow$ & Vague$\downarrow$ & Factuality$\uparrow$ & \textbf{Utility} $\uparrow$ \\
\midrule
\multicolumn{7}{l}{\textit{(a) Reward composition on} \method} \\
R1 only & 65.8 & 30.6 & 41.7 & 4.5 & 28.3 & 85.4 & 0.747 \\
R1 + R2 & 69.2 & 35.7 & 41.9 & 3.6 & 22.4 & 91.7 & 0.799 \\
R1 + R2 + R3 & \textbf{72.9} & \textbf{39.8} & 44.1 & \textbf{1.4} & \textbf{19.6} & \textbf{93.9} & \textbf{0.917} \\
\midrule
\multicolumn{7}{l}{\textit{(b) Contrastive refinement }} \\
GRPO only & 68.6 & 26.8 & 39.5 & 3.2 & 31.7 & 90.9 & 0.793 \\
+ Booster contr. only & 69.1 & 29.5 & 41.7 & 3.2 & 20.6 & 92.4 & 0.799 \\
+ Utilizer contr. only & 71.7 & 35.9 & 41.7 & 1.8 & 22.1 & 91.7 & 0.906 \\
+ Both (full \method) & \textbf{72.9} & \textbf{39.8} & 44.1 & \textbf{1.4} & \textbf{19.6} & \textbf{93.9} & \textbf{0.917} \\
\bottomrule
\end{tabular}
}
\end{table*}

To understand what contributes to the improvements made by \method, we conducted extensive ablation studies on reward compositions, contrastive refinement components, and memory types.


\textbf{Reward composition.}
Table \ref{tab:ablations}(a) validates our stage-wise process reward. Using only R1 (end-to-end) yields Accuracy of 65.8\% and Utilization of 0.747. Adding R2 (memory quality) improves Accuracy to 69.2\% and drops Vague from 28.3\% to 22.4\%. Incorporating R3 (utilization) brings the system to full performance (72.9\%, 0.917). This confirms that \textit{sparse global signals are inadequate for training intermediate memory operations}---explicit stage-wise supervision is necessary.

\Needspace{9cm}
\begin{wrapfigure}{r}{0.6\textwidth}
    \centering
    \includegraphics[width=\linewidth]{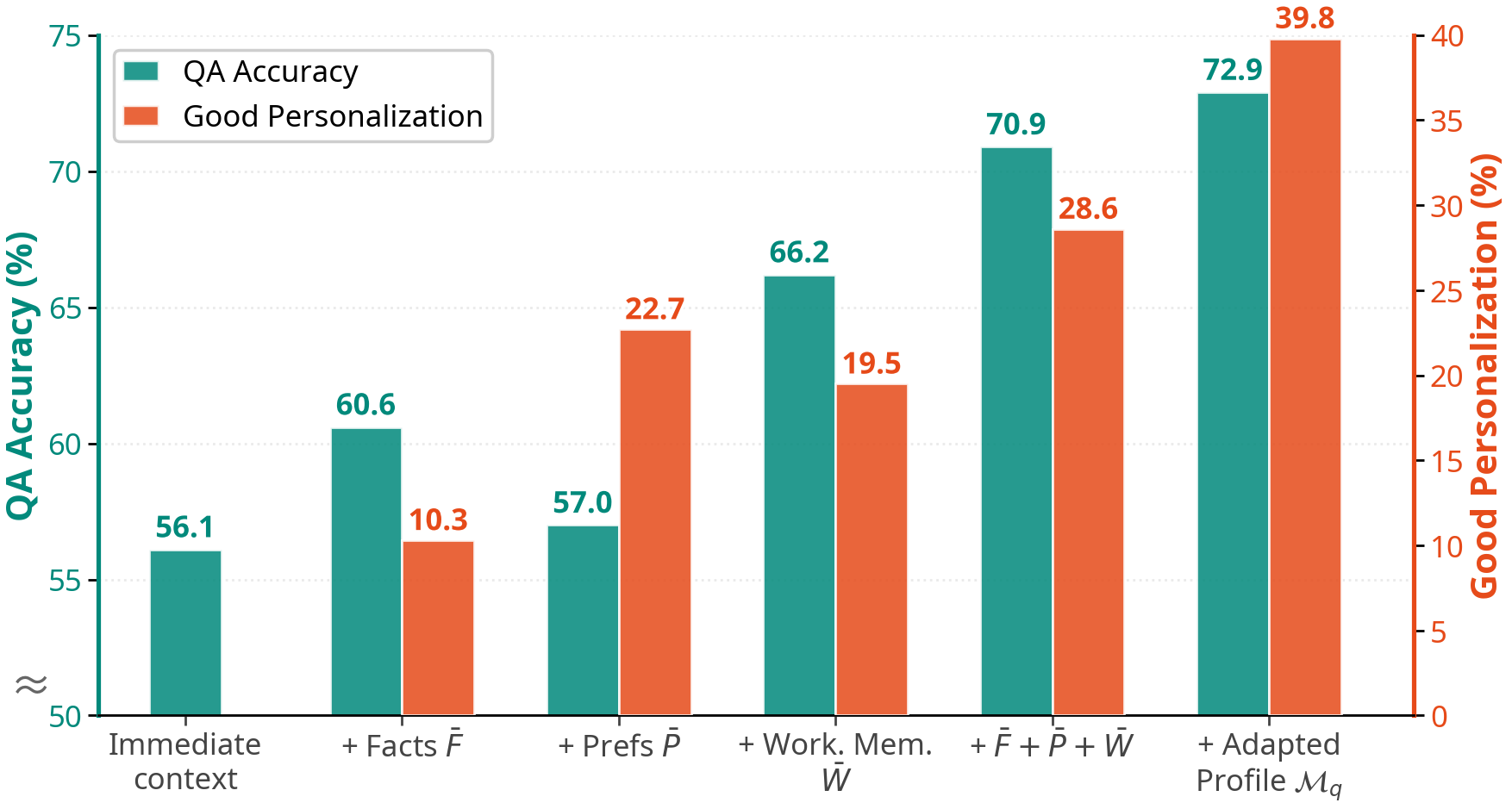}
    \caption{Impact of memory types on QA accuracy and personalization quality}
    \label{fig:memory_types}
\end{wrapfigure}
\textbf{Contrastive refinement components.}
As shown in Table \ref{tab:ablations}(b), Booster contrastive training specifically improves memory formation (raising Factuality to 92.4\%) while dropping the Vague rate from 31.7\% to 20.6\%. Conversely, Utilizer contrastive training specifically enhances retrieval (boosting Good Personalization to 35.9\% and utilization score to 0.906). Combining both ensures each module only receives gradients it earned, yielding full system performance.

\textbf{Memory type comparison.}
As illustrated in Figure \ref{fig:memory_types}, isolating the memory banks reveals their distinct cognitive roles. The Working Memory bank primarily drives QA Accuracy (66.2\%), while the Preference bank drives Good Personalization (22.7\%). Crucially, merely combining these raw banks bottlenecks Good Personalization at 28.6\%. It is only when the Utilizer synthesizes these stores into a query-adapted profile that personalization leaps to 39.8\%, proving that active context construction is strictly superior to raw retrieval.


\subsection{Training Dynamics Analysis}
\label{sec:training_dynamics}
\begin{figure}[t]
    \centering
    \includegraphics[width=\textwidth]{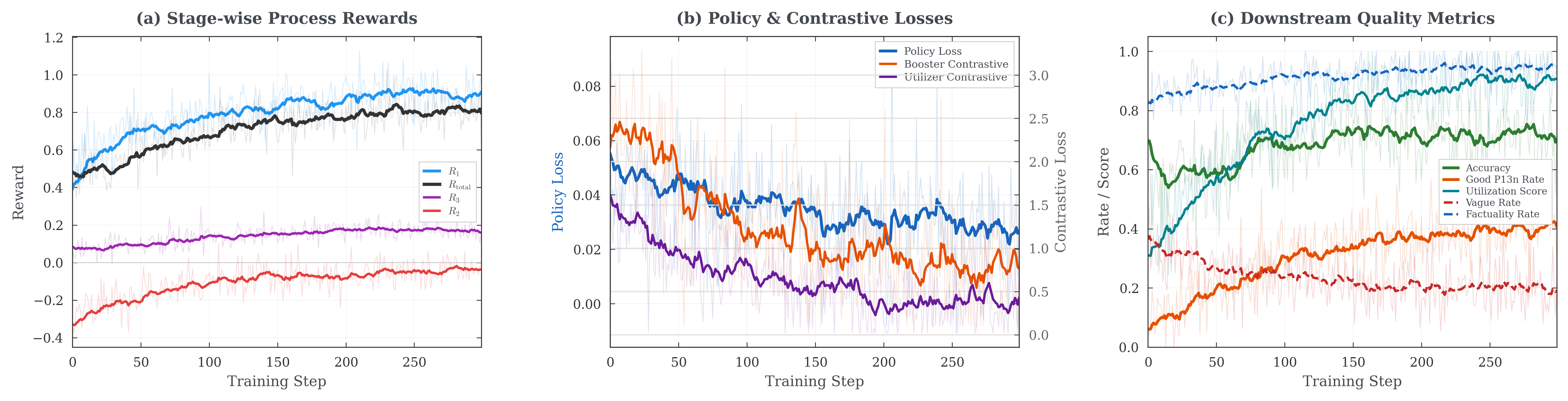}
    \caption{\textbf{Training dynamics of \method.} (a) Stage-wise process rewards. (b) Policy and contrastive losses. (c) Downstream quality metrics. Curves are EMA-smoothed.}
    \label{fig:training_dynamics}
\end{figure}

Figure~\ref{fig:training_dynamics} illustrates \method's training dynamics. Panel (a) reveals a bottom-up causal learning chain: the memory quality reward ($R_2$) improves first (rising from $-$0.26 to $-$0.04) before the utilization reward ($R_3$) and the response reward ($R_1$) follow. Panel (b) shows both contrastive losses converging (booster: 2.6$\to$0.7, utilizer: 1.4$\to$0.35), demonstrating that role-decomposed preference signals provide targeted credit assignment. Panel (c) validates that these training signals translate to downstream quality: Vague Rate drops to 19.6\%, Utilization Score reaches 0.917, and Good P13n Rate improves 6$\times$ to 39.8\%, with the improvement order mirroring the reward progression.

\subsection{Gated Error Analysis}
\label{sec:error_analysis}
We conduct a gated error analysis to isolate where failures occur in the pipeline (Figure \ref{fig:error_analysis}). Memory creation errors account for 37.9\% of failures in MemGAS and 25.3\% in Mem-R1, validating that poor memory formation is the primary bottleneck in existing systems. \method\ not only provides the lowest memory (16.4\%) and profile generation (5.8\%) error rates, but it pushes the overall system to the highest correctness rate (72.9\%). \textit{Why does \method\ succeed?} By enforcing selective attention and using the asymmetric R2 penalty, \method\ actively prevents vague or hallucinated information from polluting the memory stores. The remaining failures are largely isolated to the frozen generator after curing the upstream memory bottleneck, confirming our memory management module effectively fulfills its role.

\subsection{Efficiency \& Generalizability Analysis}
\label{sec:efficiency_generalizability_analysis}

\Needspace{9cm}
\begin{wraptable}{r}{0.6\textwidth}
\centering
\small
\caption{\textbf{(a)} Efficiency and memory store statistics; \textbf{(b)} Results on a real-world Chat AI traffic data.}
\label{tab:efficiency}
\begin{tabular}{@{}l cc@{}}
\toprule
\textbf{Metric} & \textbf{Zero-shot} & \textbf{\method} \\
\midrule
\multicolumn{3}{l}{\textit{(a) Efficiency Analysis}} \\
Avg.\ memories per user & 127.4 & 89.7  \\
Avg.\ memory token length & 8.6 & 10.9 \\
Avg.\ adapted memories per query & 1.9 & 1.7 \\
Fact : Pref : WM ratio & 21:11:68 & 9:14:77 \\
\midrule
Memory formation latency (s/episode) & 38 & 0.5 \\
Training latency (s/episode) & - & 1.14 \\
Inference latency (s/query) & 1.2 & 0.26 \\
\midrule
\midrule
\multicolumn{3}{l}{\textit{(b) Generalizability Analysis}} \\
Good P13n\% & 1.7 & 5.5 \\ 
Vague\% & 63.9 & 46.1 \\
Factuality\% & 41.9 & 60.7 \\
\bottomrule
\end{tabular}
\end{wraptable}

\textbf{Efficiency Analysis.} Beyond quality, practical deployment demands strict efficiency. 
As shown in Table \ref{tab:efficiency}(a), the Zero-shot baseline indiscriminately stores 127.4 generic memories per user, creating a severe computational bottleneck with 38 seconds of memory formation latency.
In contrast, \method's saliency filter and booster actively condense context into just 89.7 highly specific entries.
This deliberate filtering reduces memory formation latency to 0.5 seconds—a 76$\times$ speedup.
At runtime, inference latency drops nearly 5$\times$ (1.2s to 0.26s), proving that feeding the generator a synthesized profile is vastly more efficient than processing raw retrieval chunks.

\textbf{Real-World Generalizability.}
We deployed \method\ on an internal dataset of real-world Chat AI traffic.
As shown in Table \ref{tab:efficiency}(b), the relative margins remain highly consistent.
Even in this noisy, unstructured setting, \method\ more than triples the Good Personalization rate (1.7\% → 5.5\%), drops the Vague memory rate from 63.9\% to 46.1\%, and spikes memory Factuality from 41.9\% to 60.7\%.
This confirms that \method’s cognitive architecture and role-decomposed training generalize robustly beyond benchmark-specific distributions.
As shown in Table \ref{tab:efficiency}(b), \method\ consistently improves over the zero-shot baseline on a new real-world dataset, raising Factuality from 41.9\% to 60.7\% and reducing Vague rate from 63.9\% to 46.1\%, which translates to a $>3\times$ gain in Good Personalization (1.7\%$\to$5.5\%). While absolute metrics are lower---reflecting the noise of uncurated real-world interactions---the relative margins remain consistent with LoCoMo, confirming that \method\'s cognitive memory architecture and role-decomposed training generalize beyond benchmark-specific distributions.
\section{Conclusion}
\label{sec:conclusion}
We presented \method, an agentic learnable memory-management framework that addresses the fundamental challenge of conversational memory: not what to store, but how to manage it.
Our work introduces three key novelties.
First, a \textbf{cognitively-inspired memory architecture} that separates factual snapshots, subjective preferences, and working memory, enabling structured reasoning over heterogeneous user information.
Second, a \textbf{stage-wise, role-decomposed RL} training paradigm that combines hierarchical process rewards with reward-decomposed contrastive refinement, directly resolving the credit-assignment problem that has limited prior RL-based memory agents.
Third, an \textbf{extended evaluation protocol} and a new benchmark \textbf{LoCoMo-P13n} are proposed to jointly stress precise recall and nuanced personalization across diverse query types in Chat AI.
Empirically, \method\ outperforms strong baselines---while operating at 5x lower inference latency than the zero-shot baseline.
These results demonstrate that memory management, grounded in cognitive science and trained with decomposed rewards, is the key to build lifelong conversational agents.

\clearpage
\newpage
\bibliographystyle{assets/plainnat}
\bibliography{paper}

\clearpage
\newpage
\beginappendix

\appendix
\section{Use of LLM}
In paper writing, we use LLMs solely for checking typos and grammar errors; they are not used for any other purposes beyond this.
\section{Experimental Setup Details}
\subsection{Dataset Details}
\label{appendix:dataset_details}
\begin{wraptable}{r}{0.5\linewidth}
    \centering
    \caption{Dataset statistics.}
    \label{tab:locomo_stats}
    \small
    \begin{tabular}{@{}lr@{}}
    \toprule
    \textbf{Statistic} & \textbf{Value} \\
    \midrule
    Users & 28 \\
    Total QA pairs & 2,637 \\
    \quad Original (Cat 1--4) & 1,529 \\
    \quad P13n extension (Cat 6--7) & 1,108 \\
    Train / eval / test & 60 / 20 / 20\% \\
    Max turns per episode & 80 \\
    \midrule
    \multicolumn{2}{@{}l}{\textbf{Per-category counts:}} \\
    \quad Multi-hop & 282 (10.7\%) \\
    \quad Temporal & 321 (12.2\%) \\
    \quad Open-domain & 96 (3.6\%) \\
    \quad Single-hop & 830 (31.5\%) \\
    \quad Recommendation$^\dagger$ & 603 (22.9\%) \\
    \quad Implicit P13n$^\dagger$ & 505 (19.2\%) \\
    \bottomrule
    \multicolumn{2}{@{}l}
    {\footnotesize $^\dagger$Our contributed annotations.}
    \end{tabular}
\end{wraptable}
Table \ref{tab:locomo_stats} summarizes the extended LoCoMo dataset. The original benchmark~\citep{maharana2024locomo} provides QA pairs across four evaluation categories (multi-hop, temporal, open-domain, single-hop); we contribute two additional categories---recommendation and implicit personalization---that require memory-driven personalization, adding 1,108 QA pairs (42\% of the total) to form a new dataset named \textbf{LoCoMo-P13n}. Category~5 (adversarial/unanswerable) is excluded following prior work. Each user's conversation history spans multiple sessions with natural temporal gaps. 
\paragraph{How we construct the data}: we use LLM, LlaMa-3.2-70B, to semi-automatically generate the data following a 4-step strategy:(1) extracting relevant information from conversations, (2) identifying conversation reference points, (3) aggregating related information, and (4) composing questions that—rather than directly querying conversation content—utilize extracted signals to enable personalized or recommended responses. We have a linguistic expert with background in data science to do manually review on \textbf{LoCoMo-P13n}.

\paragraph{Real-world Chat AI dataset} To verify the generalizability of \method, we further collect data from real world Chat AI traffic where real users are producing conversations. We use their past 28-days chat history as the user raw history to form memory. The dataset comprises 2,719 examples spanning 69,444 sessions and 110,996 turns in total, with a median of 22 sessions (range 2--83) and 36 turns (range 5--100) per example. Sessions are short, containing 1.6 turns on average, with each turn averaging 9.8 words. Each example is additionally paired with a question-answer pair (median question/answer length of 11/31 words) and an immediate conversation context (median 8 turns, mean 18.0). Unfortunately, this dataset is not able to be open-sourced due to privacy constraint, but the results in Section \ref{sec:efficiency_generalizability_analysis} verify the effectiveness of \method\ in real-world application.

\subsection{\method\ Training Algorithm}
\method\ training steps are shown in Algorithm \ref{alg:method}.
To ensure the complex multi-stage pipeline does not bottleneck compute, our training step utilizes GPU and API resources complementarily.
First, the policy generates traces entirely on the GPU, and reference log-probabilities are pre-computed and cached.
Second, the frozen generator and reward judges run concurrently via API, allowing the reference model to be safely offloaded to the CPU.
Finally, during the update phase, only the policy model requires GPU memory since all reference probabilities were previously cached.
Each training step performs $E$ inner epochs over this cached rollout data, executing early stopping if the mean KL divergence exceeds $1.5\times$ the target threshold.
\begin{algorithm}[t]
\caption{\method: Training Step}
\label{alg:method}
\begin{algorithmic}[1]
\REQUIRE Batch of episodes $\{(\mathcal{H}_e, q_e, a^*_e)\}$, policy $\pi_\theta$, reference $\pi_{\text{ref}}$, generators/judges $\pi_{\text{gen}}, \pi_{\text{judge}}$
\FOR{each episode $e$, each rollout $g = 1, \ldots, G$}
    \STATE \textbf{Phase A (Filter):} $s_t \gets \pi_\theta^{\text{filt}}(u_t)$ for all $t$;\; $\mathcal{S} \gets \{t : s_t = \text{True}\}$
    \STATE \textbf{Phase B (Booster):} \textbf{for} $k = 1, \ldots, |\mathcal{S}|$: $\langle d_k, v_k, o_k \rangle \gets \pi_\theta^{\text{boost}}(u_{t_k}, \mathcal{M}_{k-1})$;\; $\mathcal{M}_k \gets \text{Apply}(\mathcal{M}_{k-1}, d_k, v_k, o_k)$ \hfill $\triangleright$ \textit{$t_k$: $k$-th turn in $\mathcal{S}$}
    \STATE \textbf{Phase C (Utilizer):} $\mathcal{M}_q \gets \pi_\theta^{\text{util}}(\mathcal{M}_K, q_e)$ \hfill $\triangleright$ \textit{$K = |\mathcal{S}|$}
    \STATE Cache $\log \pi_\theta^{\text{old}}(y \mid P)$ for all traces;\; pre-compute $\log \pi_{\text{ref}}(y \mid P)$
\ENDFOR
\STATE \textbf{Rewards:} $\hat{a} \gets \pi_{\text{gen}}(q_e, \mathcal{M}_q)$;\; compute $R_1, R_2, \mu$ via $\pi_{\text{judge}}$, $\kappa$ from ground-truth labels \hfill $\triangleright$ \textit{API, concurrent}
\STATE \textbf{Advantages:} $\tilde{A}_{e,g} \gets (R_{e,g} - \bar{R}_e) / \sigma_{R_e}$ for each episode $e$, where $R_{e,g} = R_1 + R_2 + R_3$ \hfill $\triangleright$ \textit{GRPO group norm.}
\STATE \textbf{Contrastive pairs:} Construct $\mathcal{P}_b$ by $R_2$ (weighted by $\Delta_{R_2}$);\; construct $\mathcal{P}_u$ by $\mu$ (weighted by $\Delta_{\mu}$), gated on $\text{sim}(\mathcal{M}^w, \mathcal{M}^l) > \tau_{\text{sim}}$
\FOR{$i = 1, \ldots, E$}
    \STATE Compute $\mathcal{L}$ via Eq.~\ref{eq:full_loss};\; update $\theta$ \hfill $\triangleright$ \textit{inner epoch}
    \IF{mean KL $> 1.5 \times$ target}
        \STATE \textbf{break} \hfill $\triangleright$ \textit{KL early stopping}
    \ENDIF
\ENDFOR
\end{algorithmic}
\end{algorithm}

\subsection{Three-Step Personalization Judge}
\label{appendix:p13n_protocol}

\begin{figure}[h]
    \centering
    \includegraphics[width=0.85\textwidth]{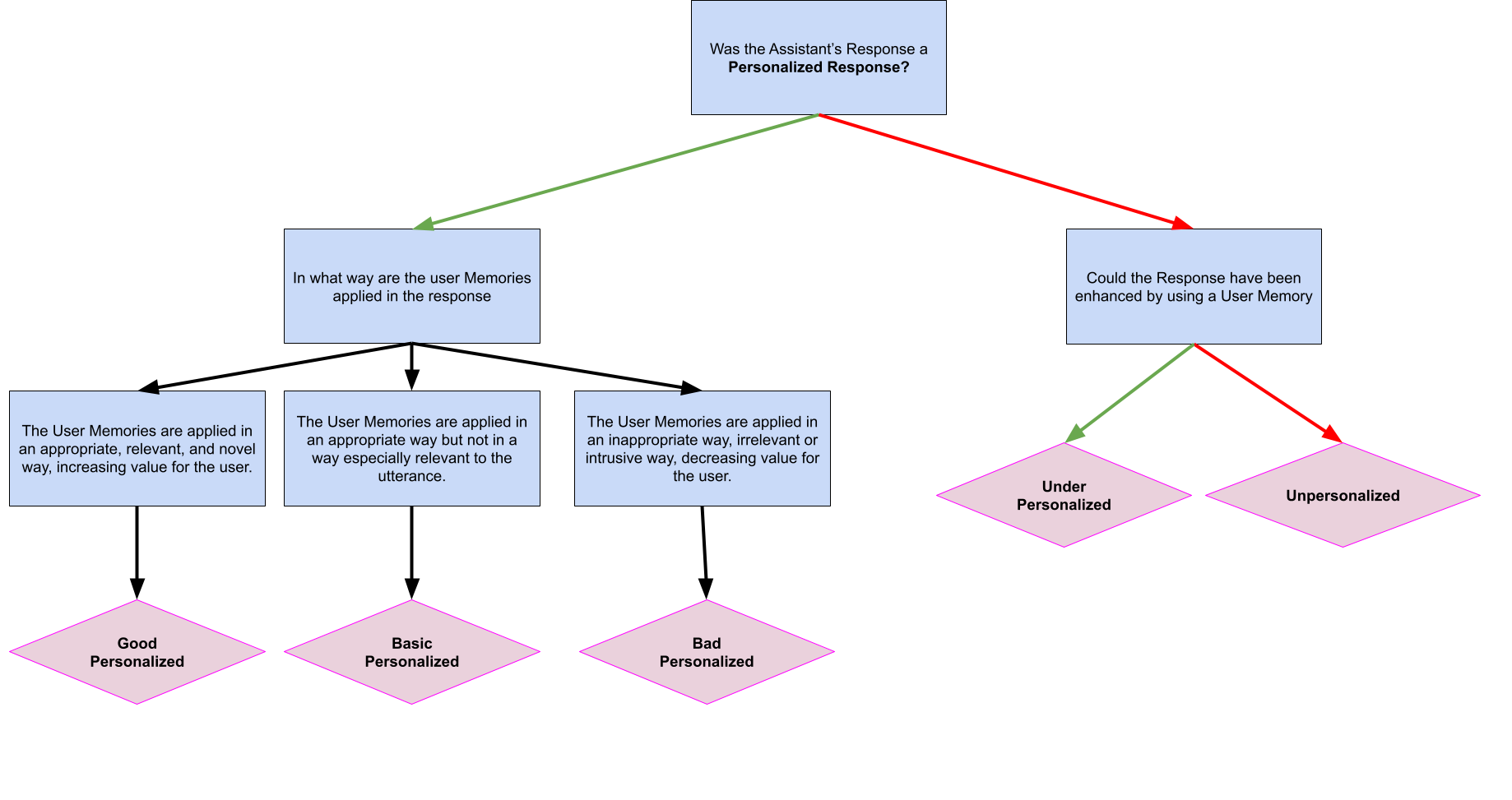}
    \caption{\textbf{Three-step personalization judge flowchart.} Task~1 detects whether the response uses memory-based personalization. If yes, Task~2 evaluates quality (Good / Basic / Bad). If no, Task~3 checks whether relevant signals existed but went unused (Underuse) or were genuinely absent (Unpersonalized). The five leaf categories are mutually exclusive.}
    \label{fig:p13n_flowchart}
\end{figure}

Standard single-score LLM-as-judge evaluation conflates whether a response \emph{uses} personal information with whether it uses it \emph{well}, and cannot detect cases where relevant memories existed but went unused. We design a three-step sequential judge that disentangles these dimensions (\ref{fig:p13n_flowchart}).

\paragraph{Task 1: Personalization Detection.}
A binary classifier determines whether the response incorporates personal information from the user's memory beyond what is already present in the query or immediate conversation. The judge considers both explicit usage (e.g., ``considering you like hiking, I recommend...'') and implicit usage (e.g., recommending a casual restaurant when the user's memory includes a preference for casual dining). Crucially, information that is already present in the user's query or conversational context does not count as personalization---the response must draw on stored memories. If personalized, proceed to Task~2; otherwise, proceed to Task~3.

\paragraph{Task 2: Personalization Quality (if personalized).}
For personalized responses, a multi-criteria judge evaluates five dimensions: (1)~\emph{richness}---does the personalization go beyond simply inserting user information, using multiple memory signals in a natural way? (2)~\emph{value-add}---does the personalization enhance the response? (3)~\emph{style}---is the tone appropriate? (4)~\emph{relevance}---is the response on-topic? (5)~\emph{quality}---is the response coherent and non-hallucinated? These combine into three quality levels:
\begin{itemize}    \item \textbf{Good}: rich personalization (criterion 1) with value-add (criterion 2) and good overall response quality (criteria 3--5).
    \item \textbf{Basic}: personalization is present but not rich or not value-adding; overall response quality is acceptable. Core case: using the user's name without deeper integration.
    \item \textbf{Bad}: personalization is misapplied (intrusive, irrelevant, or offensive) or the overall response quality is poor despite personalization.
\end{itemize}

\paragraph{Task 3: Signal Detection (if not personalized).}
For unpersonalized responses, the judge determines whether relevant memory signals \emph{existed} that could have enhanced the response but were not used. This detects \emph{underuse}---a failure mode invisible to standard evaluation. If no relevant signals exist (e.g., a purely factual query or no applicable user memory to personalize the query), the response is simply marked as not applicable (NA).

\paragraph{Five-category taxonomy.}
The three tasks yield five mutually exclusive response categories, shown in \ref{fig:p13n_flowchart}.

We derive three end-to-end personalization metrics from this taxonomy: \textbf{P13n Rate} (fraction of responses in any personalized category, including bad), \textbf{Good P13n Rate} (fraction in \textsc{Good Personalized}---the strictest measure), and \textbf{Underuse Rate} (fraction in \textsc{Underuse}---measuring missed opportunities).

\textbf{Human Agreement Test} To validate the trustworthiness of this proposed judge, two linguistic engineers from our team conducted human agreement tests on the LoCoMo-P13n and our internal datasets. We have seen an accuracy score of 91\%, 89\% and a kappa score of 0.71, 0.69 on the two datasets, respectively. The results indicate the evaluation protocol is substantially trustful for our experiments.

\subsection{Judge Cross-Validation}
\label{appendix:judge_validation}
To verify that our E2E findings are not an artifact of GPT-4o judge bias, we rerun the full E2E evaluation (QA accuracy and the three-step personalization judge) on a random subset of 200 test QA pairs from LoCoMo/LoCoMo-P13n using Claude 3.5 Sonnet as an independent judge. As shown in Table~\ref{tab:judge-crossval}, absolute scores shift only slightly between judges (within 2.5\%), with Claude marginally stricter on QA accuracy, but the method ranking is identical under both judges (Spearman's $\rho = 1.00$ for both QA accuracy and Good personalization rate), and SALIMORY remains the top method on both metrics under both judges. This confirms that the reported gains reflect genuine differences in memory quality and utilization rather than judge-specific preferences.

\begin{table}[h]
\centering
\caption{E2E results on a random 200-query subset of the LoCoMo/LoCoMo-P13n test set under two independent judges. Best in \textbf{bold}.}
\label{tab:judge-crossval}
\begin{tabular}{lcccc}
\toprule
Method & \multicolumn{2}{c}{GPT-4o} & \multicolumn{2}{c}{Claude 3.5 Sonnet} \\
\cmidrule(lr){2-3} \cmidrule(lr){4-5}
 & QA $\uparrow$ & Good $\uparrow$ & QA $\uparrow$ & Good $\uparrow$ \\
\midrule
SALIMORY & \textbf{73.4} & \textbf{40.5} & \textbf{71.8} & \textbf{38.2} \\
Zero-shot Agentic & 70.2 & 7.1 & 69.5 & 6.3 \\
Mem-R1 & 63.1 & 15.8 & 62.4 & 16.9 \\
A-Mem & 60.8 & 22.4 & 60.1 & 20.7 \\
MemoryGAS & 48.5 & 5.6 & 49.8 & 4.9 \\
Mem-$\alpha$ & 46.9 & 15.2 & 45.3 & 16.1 \\
Infinite Context & 34.1 & 17.5 & 32.8 & 18.9 \\
Mem1 & 14.6 & 2.0 & 16.1 & 1.4 \\
\bottomrule
\end{tabular}
\end{table}

\subsection{Metric Definitions}
\label{appendix:metric_details}

Table \ref{tab:metrics} provides definitions for all reported metrics.

\begin{table}[h]
\centering
\small
\caption{Evaluation metrics. E2E personalization metrics are derived from the three-step judge (\ref{appendix:p13n_protocol}); memory metrics are judged independently.}
\label{tab:metrics}
\begin{tabular}{@{}llp{7cm}@{}}
\toprule
\textbf{Metric} & \textbf{Level} & \textbf{Definition} \\
\midrule
QA Accuracy & E2E & Fraction of responses judged correct against ground truth \\
P13n Rate & E2E & Fraction in any personalized category (\ref{appendix:p13n_protocol}) \\
Good P13n Rate & E2E & Fraction in \textsc{Good Personalized} (\ref{appendix:p13n_protocol}) \\
Underuse Rate & E2E & Fraction in \textsc{Underuse} (\ref{appendix:p13n_protocol}) \\
\midrule
Vague Rate & Memory & Fraction of memory entries too generic to be actionable \\
Factuality Rate & Memory & Fraction of entries faithful to source utterances \\
Profile Util.\ Score & Memory & Judge-scored relevance of profile $\rho$ to question $q$ ($[0,1]$) \\
\midrule
Error Bucket & Diagnostic & Gated attribution of each failure to the earliest failing stage: memory $\to$ profile $\to$ generation \\
\bottomrule
\end{tabular}
\end{table}

\paragraph{Memory quality scoring.}
The judge evaluates each entry in $\mathcal{M}_K$ against the original utterance that produced it. An entry is \emph{vague} if it has lost the specificity needed to be actionable (e.g., ``User likes food'' vs.\ ``User loves Thai curry''). An entry is \emph{non-factual} if it misrepresents the source (e.g., ``thinking about getting a dog'' $\to$ ``owns a dog'').

\paragraph{Error-bucket attribution.}
For each incorrect response, we apply a gated cascade: if the profile contains sufficient information to answer correctly $\to$ \textsc{Generation Error}; else if the memory contains the information but the profile missed it $\to$ \textsc{Profile Error}; else $\to$ \textsc{Memory Error}. Each failure is assigned to the earliest failing stage.

\subsection{Hyperparameters}
\label{appendix:hyperparams}

\begin{table}[h]
\centering
\small
\caption{Hyperparameters for \method\ training.}
\label{tab:hyperparams}
\begin{tabular}{@{}llr@{}}
\toprule
\textbf{Group} & \textbf{Parameter} & \textbf{Value} \\
\midrule
\multirow{4}{*}{Training} & GRPO group size $G$ & 4 \\
& Batch size & 8 \\
& Training epochs & 5 \\
& Inner update epochs $E$ & 4 \\
\midrule
\multirow{4}{*}{Optimization} & Learning rate & $10^{-5}$ \\
& Optimizer & AdamW \\
& Warmup ratio & 0.1 \\
& Gradient clip norm & 1.0 \\
\midrule
\multirow{3}{*}{RL} & Clip ratio $\epsilon$ & 0.2 \\
& KL coefficient $\lambda_{\text{KL}}$ & 0.08 \\
& KL target (early stopping) & 1.5 \\
\midrule
\multirow{5}{*}{Reward} & Personalization weight $\lambda_p$ & 0.45 \\
& Memory penalty $\lambda_{\text{pen}}$ & 0.5 \\
& Memory bonus $\lambda_{\text{bon}}$ & 0.3 \\
& Utilization weight $\lambda_\mu$ & 0.15 \\
& Filter weight $\lambda_\kappa$ (recall $w_r{=}0.6$) & 0.1 \\
\midrule
\multirow{4}{*}{Contrastive} & Booster weight $\lambda_b$ & 0.15 \\
& Utilizer weight $\lambda_u$ & 0.1 \\
& Temperature $\eta$ & 0.15 \\
& Memory similarity threshold $\tau_{\text{sim}}$ & 0.15 \\
\bottomrule
\end{tabular}
\end{table}
For training, we use transformers and PyTorch to train on a computation node with 8 A100 GPUs.
\section{Limitation}
\label{appendix:limitation}
\Needspace{9cm}
\begin{wrapfigure}{r}{0.6\textwidth}
    \centering
    \includegraphics[width=\linewidth]{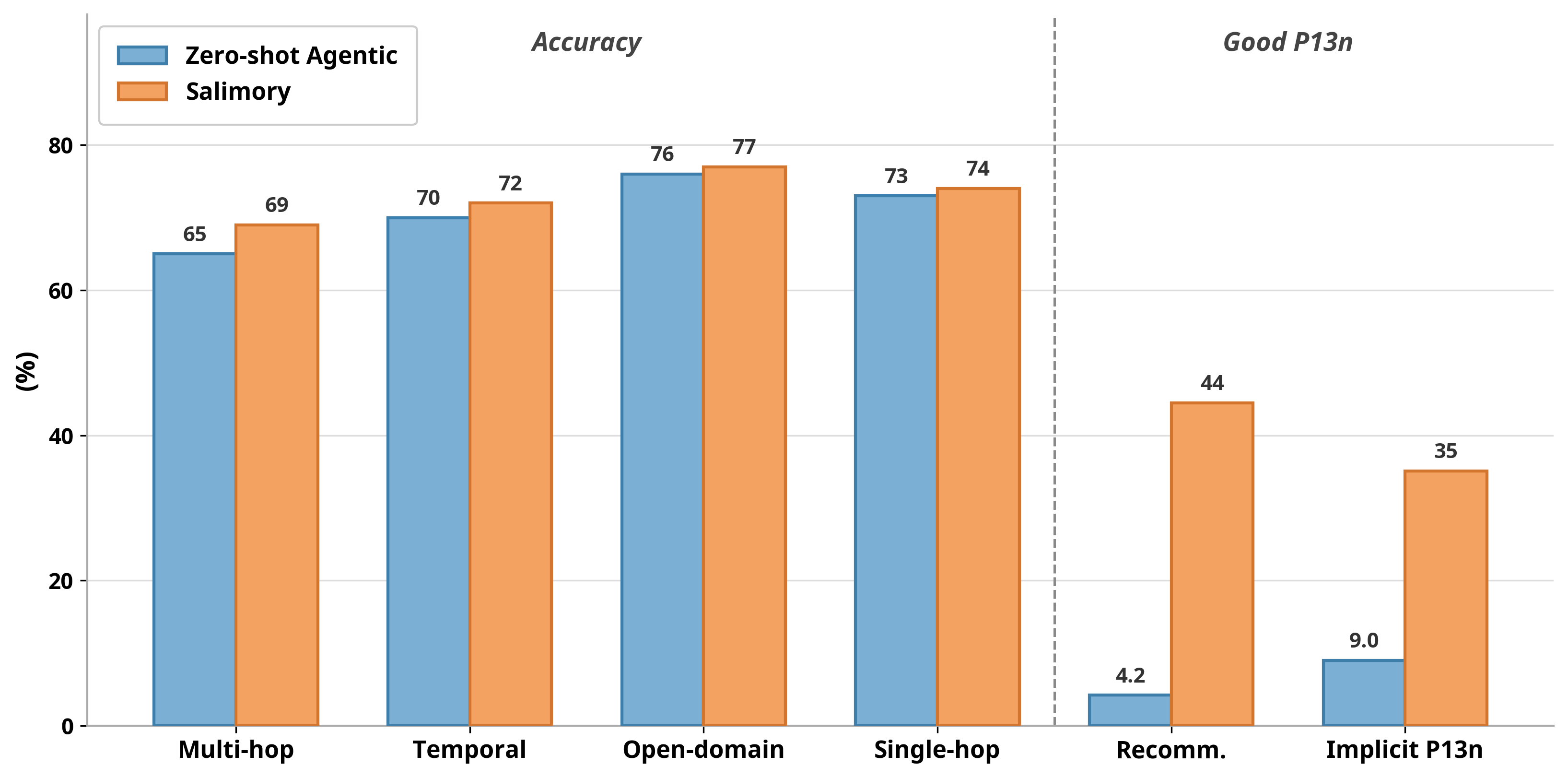}
    \caption{\textbf{Per-category accuracy.} \method\ improves all categories, with the largest gains on recommendation and multi-hop questions that require memory synthesis.} 
    \label{fig:per_category}
\end{wrapfigure}
Despite the success brought by \method, we still notice two limitations to be further studied in the future: \textbf{Dependence on a frozen large LLM for judging.} All three reward levels rely on a frozen large LLM (GPT-4o) — response accuracy, memory quality, and utilization are all assessed via LLM-as-a-judge calls. This means the training signal is bounded by the judge model's own evaluation capability and biases and further, the api calling would hinder the training stability. \textbf{Sequential memory formation limits scalability to long conversations.} The booster must process salient turns sequentially (each turn needs the updated memory state from the previous one), meaning memory formation latency scales linearly with conversation length. For the current dataset (~16 turns/episode), this is manageable, but for real-world deployment with hundreds or thousands of turns per user, this becomes a bottleneck that batch parallelization cannot resolve.
\section{Supplementary Analysis}
\label{appendix:supplementary_analysis}
\subsection{Comparative Results on LongMemEval-s}
\begin{table*}[t]
\centering
\small
\caption{\textbf{Main results on LongMemEval-s.} Response quality and memory quality are reported. Best in \textbf{bold}. $\uparrow$/$\downarrow$: higher/lower is better.}
\label{tab:main_results_s}
\resizebox{\textwidth}{!}{%
\begin{tabular}{@{}l >{\columncolor{orange!15}}c >{\columncolor{orange!15}}c cccc cc@{}}
\toprule
& \multicolumn{6}{c}{\textbf{Response Quality}} & \multicolumn{2}{c}{\textbf{Memory}} \\
\cmidrule(lr){2-7}
& QA & \multicolumn{5}{c}{Personalization Categories (\%)} & \multicolumn{2}{c}{\textbf{Quality}} \\
\cmidrule(lr){2-2} \cmidrule(lr){3-7} \cmidrule(lr){8-9}
\textbf{Method} & Acc.\ $\uparrow$ & Good $\uparrow$ & Basic & Bad $\downarrow$ & Under.\ $\downarrow$ & NA\ $\downarrow$ & Vague $\downarrow$ & Fact.\ $\uparrow$ \\
\midrule
\multicolumn{9}{l}{(b) Comparative Results on \textbf{\textit{LongMemEval-s}}} \\
Infinite Context & 47.7 & 9.6 & 22.8 & 5.7 & 42.9 & 19.0 & - & - \\
Zero-shot Agentic & 70.9 & 8.0 & 27.3 & 6.8 & 40.6 & 17.3 & 25.5 & 89.1 \\
A-Mem & 60.0 & 13.6 & 22.9 & 1.7 & 12.5 & 49.3 & 21.1 & 53.7 \\
MemoryGAS & 60.2 & 7.6 & 18.1 & 2.0 & 19.2 & 53.1 & 19.9 & 58.1 \\
Mem1 & 51.4 & 14.6 & 31.6 & 2.9 & 18.0 & 32.9 & 15.9 & 81.6 \\
Mem-$\alpha$ & 54.8 & 13.8 & 32.2 & 2.2 & 19.0 & 32.8 & 9.6 & 82.2 \\
Mem-R1 & 66.7 & 15.1 & 37.2 & 4.5 & 29.3 & 13.9 & 9.9 & 82.8 \\
\midrule
\method\ & \textbf{69.1} &\textbf{20.8} & \textbf{43.7} & \textbf{0.8} & \textbf{6.8} & \textbf{27.9} & \textbf{6.3} & \textbf{93.3} \\
\bottomrule
\end{tabular}
}
\end{table*}

\subsection{Backbone model and Training strategy ablation}
\begin{table*}[t]
\centering
\setlength{\tabcolsep}{5pt}
\renewcommand{\arraystretch}{0.95}
\caption{\method\ results using different training strategies and backbone models. Best in each section in \textbf{bold}. $\uparrow$/$\downarrow$: higher/lower is better.}
\label{tab:supplementary_ablations}
\begin{tabular}{@{}lcccccc@{}}
\toprule
& \multicolumn{3}{c}{\textbf{E2E Quality}} & \multicolumn{2}{c}{\textbf{Memory Quality}} & \textbf{Memory Utilization} \\
\cmidrule(lr){2-4} \cmidrule(lr){5-6} \cmidrule(lr){7-7}
\textbf{Config} & Acc.$\uparrow$ & Basic P13n & Good P13n$\uparrow$ & Vague$\downarrow$ & Fact.$\uparrow$ & Util.$\uparrow$ \\
\midrule
\multicolumn{7}{l}{\textit{(a) Training strategy}} \\
REINFORCE & 36.9 & 23.7 & 12.9 & 68.6 & 43.9 & 0.291 \\
PPO & 39.2 & 23.9 & 15.7 & 67.9 & 48.6 & 0.393 \\
GRPO & 68.6 & 39.5 & 26.8 & 31.7 & 90.9 & 0.793 \\
\method\ & \textbf{72.9} & 44.1 & \textbf{39.8} & \textbf{19.6} & \textbf{93.9} & \textbf{0.917} \\
\midrule
\multicolumn{7}{l}{\textit{(b) Backbone models}} \\
Llama3.2-1B & 29.6 & 9.7 & 0.9 & 74.1 & 40.3 & 0.148 \\
Llama3.2-3B & 58.8 & 19.4 & 11.7 & 39.1 & 71.5 & 0.476 \\
Qwen3.5-9B & \textbf{72.9} & 44.1 & \textbf{39.8} & \textbf{19.6} & \textbf{93.9} & \textbf{0.917} \\
\bottomrule
\end{tabular}
\end{table*}
\paragraph{Backbone models.} Table \ref{tab:supplementary_ablations}(b) shows that Llama3.2-1B collapses (Accuracy 29.6\%, Vague 74.1\%), Llama3.2-3B achieves 58.8\% Accuracy, and Qwen3.5-9B reaches the optimal 72.9\%. This suggests that the cognitive memory roles require a minimum threshold of base reasoning capacity to execute effectively. We probe the base policy with 200 canary prefixes from LoCoMo test conversations; verbatim continuations occur in 0\% of cases, indicating no measurable memorization.
\paragraph{Training strategy.} As presented in Table \ref{tab:ablations}(a), REINFORCE and PPO yield Accuracy of only 36.9\% and 39.2\%, with Vague rates remaining high at 68.6\% and 67.9\%. Switching to GRPO provides a massive leap (Accuracy 68.6\%, Vague 31.7\%), indicating that relative scoring within a batch is crucial for stable learning in long-horizon memory tasks. \textit{Why does \method\ push further?} By augmenting GRPO with contrastive refinement, \method\ achieves 72.9\% Accuracy and 19.6\% Vague rate, verifying that standard GRPO alone is insufficient to fully resolve credit assignment in multi-stage memory management.

\subsection{Frozen-Generator Ablation}
Our main results use a frozen Qwen3-235B-A22B-Instruct generator, raising the question of whether the gains stem from generator capacity rather than the learned memory module. We rerun E2E evaluation with the trained SALIMORY policy held fixed and the generator swapped to Qwen3.5-9B-Instruct, matching the policy's scale. As shown in Table~\ref{tab:generator-ablation}, absolute scores drop for all methods under the weaker generator, as expected, but SALIMORY retains the lead on both QA accuracy and Good personalization, with its QA margin over Mem-R1 widening from 10.2pp to 12.5pp. Memory-quality metrics (Vague, Factuality) are unaffected by construction, since they are judged on the policy-produced memory bank. This confirms that the improvements originate in the memory-management module rather than in generator scale---high-quality, well-utilized memories matter more, not less, when the generator is weaker.

\begin{table}[h]
\centering
\caption{E2E results with the default large generator versus the policy-matched small generator on LoCoMo/LoCoMo-P13n. Best in \textbf{bold}.}
\label{tab:generator-ablation}
\begin{tabular}{lcccc}
\toprule
Method & \multicolumn{2}{c}{Qwen3-235B-A22B} & \multicolumn{2}{c}{Qwen3.5-9B} \\
\cmidrule(lr){2-3} \cmidrule(lr){4-5}
 & QA $\uparrow$ & Good $\uparrow$ & QA $\uparrow$ & Good $\uparrow$ \\
\midrule
SALIMORY & \textbf{72.9} & \textbf{39.8} & \textbf{64.8} & \textbf{34.2} \\
Zero-shot Agentic & 70.9 & 6.6 & 58.6 & 4.8 \\
Mem-R1 & 62.7 & 16.3 & 52.3 & 12.1 \\
A-Mem & 61.6 & 21.9 & 50.2 & 17.3 \\
\bottomrule
\end{tabular}
\end{table}

\subsection{Memory Comparison Analysis}
\label{appendix:memory_comparison}

\begin{table*}
\centering
\caption{\textbf{Impact of vague memories.} Filtering vague entries nearly doubles good-personalization rate.}
\label{tab:vague_impact}
\begin{tabular}{@{}l cccc@{}}
\toprule
\textbf{Memory Condition} & Recall@5$\uparrow$ & Win Rate$\uparrow$ & P13n$\uparrow$ & Good P13n$\uparrow$ \\
\midrule
With vague memories    & 35.9 & 34.6 & 29.6 & 22.2 \\
Without vague memories & \textbf{52.1} & \textbf{40.9} & \textbf{49.3} & \textbf{40.9} \\
\bottomrule
\end{tabular}
\end{table*}
We directly validate that memory quality drives downstream performance by comparing retrieval under controlled memory conditions (\ref{tab:vague_impact}).
Filtering vague entries improves Recall@5 from 35.9\% to 52.1\% and good-personalization from 22.2\% to 40.9\%. Vague memories act as retrieval noise, displacing useful entries from top-$k$ results. \method's booster directly addresses this via the asymmetric $R_2$ penalty.

\subsection{More Memory $\neq$ Better Responses}
\label{appendix:analysis_infinite}
\begin{table*}
\centering
\caption{\textbf{Infinite context degrades performance.} More memory increases personalization \emph{attempts} but collapses their quality.}
\label{tab:infinite_context}
\begin{tabular}{@{}l cccc@{}}
\toprule
\textbf{Context} & Win Rate$\uparrow$ & P13n$\uparrow$ & Good P13n$\uparrow$ & Mis-rec$\downarrow$ \\
\midrule
Top-5 retrieval   & 40.9 & 49.3 & 40.9 & 8.3 \\
Infinite context  & 27.6 & 52.6 & 22.7 & 29.6 \\
\bottomrule
\end{tabular}
\end{table*}
Table \ref{tab:infinite_context} tests whether providing all memories (infinite context) yields an upper bound on performance.
Win rate drops from 40.9\% to 27.6\% and mis-recommendation rate spikes from 8.3\% to 29.6\%. The generator attempts \emph{more} personalization (P13n rate rises) but gets it wrong (good P13n collapses). This motivates the utilizer's role: compress and filter memories into a question-relevant profile rather than passing raw context.

\subsection{Per-Category Breakdown}
Figure \ref{fig:per_category} reports performance by question category. \method\ achieves the largest accuracy gains on multi-hop (+4pp) and recommendation (+5pp)---categories requiring multi-entry synthesis---while single-hop and open-domain show minimal improvement (+1pp), as single-fact retrieval already works well. For personalization, \method\ lifts Good P13n from 4.2\% to 44.5\% on recommendation and from 9.0\% to 35.1\% on implicit personalization, confirming that our cognitive memory architecture and trained utilizer generalize most strongly to synthesis-heavy queries.

\section{Reward Judge Prompts}
\label{appendix:judge_prompts}

All reward signals in \method\ are computed by frozen LLM judges. Below we provide the judge prompt templates for each reward level.

\subsection{R1 Response Quality Judge}
\label{appendix:l1_prompt}

\begin{tcolorbox}[colback=gray!5, colframe=gray!60, title=R1 Response Quality, fonttitle=\small\bfseries]
\tiny
\#\# Context
You are a judge who evaluates a chatbot response on two independent axes: accuracy and personalization.

The user has the following personal information:
Memory: {memory}

Here is the immediately preceding conversation:
{conversation}

Here is the user query:
{question}

Here is the response to evaluate:
{response}

Here is the golden response (ground truth):
{label}

\#\# Evaluation Criteria

\#\#\# 1. Correctness (boolean)
Does the response answer the query correctly? Compare against the golden response.
- If the response is factually wrong or fails to answer the question, correctness is false.
- If the response addresses the question correctly (even if worded differently from the golden response), correctness is true.

\#\#\# 2. Accuracy Score (0.0–1.0)
How correct is the response compared to the golden answer? This is independent of personalization.
- 0.0: Completely wrong or fails to answer the question.
- 0.3: Partially addresses the question but with significant errors.
- 0.5: Partially correct with some relevant information.
- 0.7: Mostly correct with minor gaps.
- 1.0: Fully correct, equivalent to the golden response.

\#\#\# 3. Personalization Score (0.0–1.0 or "n/a")
How well does the response use personal information from memory?
- 0.0: Misused or irrelevant personal information that hurts the response.
- 0.3: Personal info acknowledged but not meaningfully integrated.
- 0.5: Superficial use of personal information.
- 0.7: Good use of relevant personal information that enhances the response.
- 1.0: Excellent use of personal information, comparable to the golden response.
- "n/a": The question does not require personalization, or no relevant memory exists.

\#\# Output
Provide your evaluation as a JSON object with:
- `correctness`: Boolean (true/false).
- `accuracy\_score`: String, a number from 0.0 to 1.0.
- `personalization\_score`: String, a number from 0.0 to 1.0, or "n/a" if not applicable.
Output the JSON object only, nothing else.
\end{tcolorbox}

\subsection{R2 Memory Quality Judge}
\label{appendix:l2_prompt}

\begin{tcolorbox}[colback=gray!5, colframe=gray!60, title=R2 Memory Quality, fonttitle=\small\bfseries]
\tiny
\#\#\# Task Description
You are a professional memory analyst. Your job is to evaluate a memory collection extracted from a conversation on two axes: vagueness and factuality.

\#\#\# Key Definitions

\#\#\#\# Vagueness
- **Vague memory entry**: A memory entry that is too generic to be actionable for personalization. It loses specific details from the original conversation. For example, "User likes food" is vague because it lost what kind of food; "User traveled" is vague because it lost where and when.
- **Specific memory entry**: A memory entry that captures concrete, actionable information that can personalize future responses. For example, "User has a peanut allergy" or "User returned from Japan trip last week" are specific.

\#\#\#\# Factuality
- **Factual memory entry**: A memory entry that accurately represents what was said in the conversation, without hallucinating or distorting details.
- **Non-factual memory entry**: A memory entry that misrepresents what was said, invents details, or draws incorrect conclusions.

\#\#\# Vagueness Examples

Conversation utterances:
- Can you recommend some Thai restaurants near downtown?
- I just got back from a trip to Japan last week
- I'm training for a marathon in October

Vague memory: {{"user\_facts": ["User likes food", "User is interested in running"], "working\_memory": ["User traveled recently"]}}
-> 3 entries, all 3 are vague -> vague\_rate = 1.0

Specific memory: {{"user\_facts": ["User has a peanut allergy", "User is training for a marathon in October"], "working\_memory": ["User returned from Japan trip last week", "User is looking for Thai restaurants near downtown"]}}
-> 4 entries, 0 are vague -> vague\_rate = 0.0

\#\#\# Factuality Examples

Utterance: "I went to Japan last week" -> Memory: "User went to China last week" -> **Non-factual** (wrong country)
Utterance: "I'm thinking about getting a dog" -> Memory: "User owns a dog" -> **Non-factual** (considering is not owning)
Utterance: "My daughter starts college in September" -> Memory: "User's daughter is starting college in September" -> **Factual**

\#\#\# Input
Conversation utterances:
{utterances}

Extracted memory:
{memory}

\#\#\# Instructions
1. List all individual memory entries across all categories (user\_facts, user\_preference, working\_memory, etc.).
2. For each entry, determine if it is vague or specific. Calculate: vague\_rate = (number of vague entries) / (total number of entries). If there are no entries, vague\_rate = 0.0.
3. For each entry, determine if it is factual or non-factual based on the conversation. Calculate: factuality\_rate = (number of factual entries) / (total number of entries). If there are no entries, factuality\_rate = 1.0.

\#\#\# Output
Output a JSON object with:
- `vague\_rate`: A string representing a number from 0.0 to 1.0, indicating the fraction of memory entries that are vague.
- `factuality\_rate`: A string representing a number from 0.0 to 1.0, indicating the fraction of memory entries that are factual.
Output the JSON object only, nothing else.
\end{tcolorbox}

\subsection{R3 Utilization Quality Judge}
\label{appendix:l3_prompt}

\begin{tcolorbox}[colback=gray!5, colframe=gray!60, title=R3 Utilization Quality, fonttitle=\small\bfseries]
\tiny
\#\#\# Task Description
You are a professional analyst evaluating the quality of a user profile summary for answering a specific question.

\#\#\# Context
A utilizer agent has created a user profile summary from memory and conversation to help answer a question. Your job is to evaluate how useful this profile is for generating a correct, personalized answer.

\#\#\# Input
Question: {question}

User Profile Summary: {user\_profile}

Memory State: {memory\_state}

Conversation: {conversation}

Golden Answer (ground truth): {ground\_truth}

\#\#\# Evaluation Rubric (0.0–1.0)
- 0.0: Profile is irrelevant or contains wrong information that would mislead the answer.
- 0.3: Profile exists but misses key information needed to answer the question.
- 0.5: Contains some relevant information but incomplete or noisy.
- 0.7: Captures the essential information needed for answering the question correctly.
- 1.0: Highly relevant, accurate, and provides exactly the right context for a personalized answer.

\#\#\# Output
Output a JSON object with:
- `utilization\_score`: A string representing a number from 0.0 to 1.0.
Output the JSON object only, nothing else.
\end{tcolorbox}

\end{document}